\documentclass{acm_proc_article-sp}

\usepackage{latexsym}
\usepackage{amsmath}
\usepackage{multirow}
\usepackage{url}
\usepackage{graphics}

\title{Identifying Purpose Behind Electoral Tweets}

\numberofauthors{2}
 \author{Saif M. Mohammad, Svetlana Kiritchenko, and Joel Martin\\
  	National Research Council Canada\\
 	 Ottawa, Ontario, Canada K1A 0R6\\
   {\tt \{saif.mohammad,svetlana.kiritchenko,joel.martin\}@nrc-cnrc.gc.ca} }

\date{}

\begin{document}
\maketitle
\begin{abstract}
Tweets pertaining to a single event, such as a national election, can number in the hundreds of millions.
Automatically analyzing them is beneficial in many downstream natural language applications such as question answering and summarization.
In this paper, we propose a new task: identifying the purpose behind electoral tweets---why do people post election-oriented tweets?
We show that identifying purpose is correlated with the related phenomenon of sentiment and emotion detection, but yet significantly different. 
 Detecting purpose has a number of applications including detecting the mood of the electorate, estimating the popularity of policies,
 identifying key issues of contention, and predicting the course of events.
We create a large dataset of electoral tweets and annotate a few thousand tweets for purpose.
We develop a system that automatically classifies electoral tweets as per their purpose,
obtaining an accuracy of 43.56\% on an 11-class task and an accuracy of 73.91\% on a 3-class task (both accuracies
well above the most-frequent-class baseline).
Finally, we show that resources developed for emotion detection
are also helpful for detecting purpose. 
% In the process, we create a new word--emotion association resource pertaining to hundreds of emotions.
\end{abstract}

\section{Introduction}

The number of tweets pertaining to a single event or topic such as a national election, a natural disaster, or gun control laws, can grow to the hundreds of millions.
The large number of tweets negates the possibility of a single person reading all of them to gain an overall global perspective.
Thus, automatically analyzing tweets is beneficial in many downstream natural language applications such as question answering and summarization.

An important facet in understanding tweets is the question of `Why?', that is, what is the purpose of the tweet?
There has been some prior work in this regard \cite{Alhadi11,Naaman10,Sankaranarayanan09}, however, they have focused on the general motivations and reasons for tweeting.
For example, Naaman et al.\@ \shortcite{Naaman10} proposed the categories of: 
information sharing, self promotion, opinions,
statements, me now, questions, presence maintenance, anecdote (me), and anecdote
(others).
On the other hand, the dominant reasons for tweeting vary when tweeting about specific topics and events. For example, the
reasons for tweeting in national elections are very different from the reasons for tweeting during a natural disaster, such as an earthquake.

There is growing interest in analyzing political tweets in particular because of a number of applications such as
determining political alignment of tweeters \cite{Golbeck11,Conover11b}, 
identifying contentious issues and political opinions \cite{Maynard11}, 
detecting the amount of polarization in the electorate \cite{Conover11},
and so on.
There is even a body of work claiming that analyzing political tweets can help predict the outcome of elections \cite{Bermingham11,Tumasjan10}.
However, that claim is questioned by more recent work \cite{Avello12}.

In this paper, we propose the task of identifying the purpose behind electoral tweets.
For example, some tweets are meant to criticize, some to praise, some to express disagreement, and so on.
Determining the purpose behind electoral tweets can help many applications such as those listed above.
There are many reasons why people criticize, praise, etc, but that is beyond the scope of this paper.
For discussions on user satisfaction from tweets we refer the reader to Liu, Cheung, and Lee \shortcite{Liu10} and Cheung and Lee \shortcite{Cheung09}.

First, we automatically compile a dataset of electoral tweets using a few hand-chosen hashtags.
We choose the 2012 US presidential elections as our target domain.
We develop a questionnaire to annotate tweets for purpose by crowdsourcing.
We analyze the annotations to determine the distributions of different kinds of purpose.
We develop a preliminary system that automatically classifies electoral tweets as per their purpose,
using various features that have traditionally been used in tweet classification, such as word ngrams and elongated words, as well as features pertaining
to eight basic emotions.
We show that resources developed for emotion detection are also helpful for detecting purpose. 

We then add to this system features pertaining to hundreds of fine emotion categories.
We show that these features lead to significant improvements in accuracy above and beyond those obtained
by the competitive preliminary system.
The system obtains an accuracy of 43.56\% on a 11-class task and an accuracy of 73.91\% on a 3-class task.
% (both accuracies well above the most-frequent-class baseline).

% We also show that purpose is correlated with sentiment and emotion. 
% In the process, we create a new emotion lexicon pertaining to hundreds of emotions that is also potentially useful
% in a number of affect analysis tasks.
Finally, we show that emotion detection alone can fail to distinguish
between several different types of purpose. For example, the same emotion of disgust
can be associated with many different kinds of purpose such as `to criticize', `to vent', and `to ridicule'.
Thus, detecting purpose provides information that is not provided simply by detecting sentiment or emotion.
We publicly release all the data created as part of this project: about 1 million original tweets on the 2012 US elections, about 2,000 tweets annotated for purpose,
about 1,200 tweets annotated for emotion, and the new emotion lexicon.\footnote{Email Saif Mohammad: saif.mohammad@nrc-cnrc.gc.ca.}

We begin with related work (Section 2). We then describe how we collected (Section 3.1) and annotated the data (Section 3.2). Section 3.3
gives an analysis of the annotations including distributions of various kinds of purpose, inter-annotator agreement, and confusion matrices.
In Section 3.4, we tease our the partial correlation and the distinction between purpose and affect.
In Section 4, we present first a basic system to classify tweets by purpose (Section 4.1), and then we describe
how we created an emotion resource pertaining to hundreds of emotions and used it to further improve performance of the basic system (Section 4.2)
In Section 5, we discuss some of the findings of the automatic classifiers and also further delineate the relation between
purpose detection and emotion detection. We present concluding remarks in Section 6.

\section{Related Work}

There exists considerable work on tweet classification by topic \cite{sakaki2010earthquake,lee2011twitter,nishida2011tweet}.
Some of the classification work that comes close to identifying purpose is described below.
Alhadi et al.\@ \shortcite{Alhadi11} annotated 1000 tweets into the categories of social interaction with people,
promotion or marketing, share resources, give or require feedback, broadcast alert/urgent information,
require/raise funding, recruit worker, and express emotions.
Naaman et al.\@ \shortcite{Naaman10} organized 3379 tweets into the categories of information sharing, self promotion, opinions,
statements, me now, questions, presence maintenance, anecdote (me), and anecdote
(others).
\cite{Sankaranarayanan09} built a system to identify tweets pertaining to breaking news.
Sriram et al.\@ \shortcite{Sriram10} annotated 5407 tweets into news,
events, opinions, deals and private messages.

Tweet categorization work within a particular domain includes
that by Collier, Son, and Nguyen \shortcite{Nigel11}, where flu-related tweets were classified into
avoidance behavior, increased sanitation, seeking pharmaceutical intervention, wearing a mask, and self reported diagnosis, and 
work by Caragea et al.\@ \shortcite{Caragea11}, where earthquake-related tweets were classified into  medical emergency, people trapped,
food shortage, water shortage, water sanitation, shelter needed, collapsed structure, food
distribution, hospital/clinic services, and person news.

To the best of our knowledge, there is no work yet on classifying electoral or political tweets into sub-categories.
As mentioned earlier, there exists work on determining political alignment of tweeters \cite{Golbeck11,Conover11b},
identifying contentious issues and political opinions \cite{Maynard11},
detecting the amount of polarization in the electorate \cite{Conover11}, and 
detecting sentiment in political tweets \cite{Bermingham11,ChungM11,OConnor10}.

Sentiment classification of general (non-domain) tweets has received much attention \cite{pak2010twitter,Jiang11,kouloumpis2011twitter}. Beyond simply positive and negative sentiment, some recent work also classifies tweets
into emotions \cite{Kim12,Mohammad12,Quercia12,Tsagkalidou11}.
 Much of this work focused on emotions argued to be the most basic.
  For example, Ekman\@ \shortcite{Ekman92} proposed six basic emotions---joy, sadness,
   anger, fear, disgust, and surprise. 
Plutchik\@ \shortcite{Plutchik80} argued in favor of eight---Ekman's six, trust, and anticipation.  There is less work on complex
   emotions, such as work by Pearl and Steyvers\@ \shortcite{PearlS10} that focused on
   politeness, rudeness, embarrassment, formality, persuasion, deception,
   confidence, and disbelief.  

Many of the automatic emotion classification systems use affect lexicons
such as the NRC emotion lexicon \cite{MohammadT10}, WordNet Affect \cite{StrapparavaV04}, and the
Affective Norms for English Words.\footnote{http://csea.phhp.ufl.edu/media/anewmessage.html}
Affect lexicons are lists of words and associated emotions and sentiments.
We will show that affect lexicons are helpful for detecting purpose behind tweets as well.

\section{Data Collection and Annotation of Purpose}
In the subsections below we describe how we collected tweets posted during the run up to the 2012 US presidential elections
and how we annotated them for purpose by crowdsourcing.

\begin{table}[t]
\caption{\label{tab:poltags} Query terms used to collect tweets pertaining to the  2012 US presidential elections.}
\begin{center}
% {\small
\resizebox{0.47\textwidth}{!}{
 \begin{tabular}{lll}
% \hline \bf Category & \bf \# of instances \\ \hline
\hline
\#4moreyears
&\#Barack
&\#campaign2012 \\
\#dems2012
&\#democrats
&\#election\\
\#election2012
&\#gop2012
&\#gop\\
\#joebiden2012
&\#mitt2012
&\#Obama\\
\#ObamaBiden2012
&\#PaulRyan2012
&\#president\\
\#president2012
&\#Romney
&\#republicans\\
\#RomneyRyan2012
&\#veep2012
&\#VP2012\\
Barack
&Obama
&Romney\\
\hline
\end{tabular}
}
\end{center}
\end{table}

\subsection{Identifying Electoral Tweets} 
We created a corpus of tweets by polling the Twitter Search API, during August and September 2012, for tweets that contained commonly known
hashtags pertaining to the 2012 US presidential elections. Table \ref{tab:poltags} shows the query terms
we used. Apart from 21 hashtags, we also collected tweets with the words Obama, Barack, or Romney. We used
these additional terms because they were the names of the two presidential candidates. 
Further, the probability that these words were used to refer to someone other than the presidential candidates was low.

The Twitter Search API was polled every four hours to obtain new tweets that matched the query.
Close to one million tweets were collected, which we will make freely available to the research community. Note that Twitter
imposes restrictions on direct distribution of tweets, but allows the distribution of tweet ids. 
One may download tweets using tweet ids and third party tools, provided those tweets have not been deleted by the people who posted them.
The query terms which produced the highest number of tweets were those involving the names of the
presidential candidates, as well as \#election2012, \#campaign, \#gop, and \#president.

We used the metadata tag ``iso\_language\_code'' to identify English tweets. Since this
tag does not always correctly reflect the language of the tweet, we also discarded tweets
that did not have at least two valid English words. We used the Roget Thesaurus
as the English word inventory. This step also helps discard very short tweets and tweets with a large proportion of misspelled words.

 Since we were interested in determining the purpose behind the tweets, we decided to focus
 on original tweets as opposed to retweets. Retweets can easily be identified through the presence of RT, rt, or Rt
 in the tweet (usually in the beginning of the post). All such tweets were discarded.

% A histogram of the number of tweets per person is shown in Table \ref{tab:histo}.
%   (Only part of the histogram is shown due to space constraints).
% The maximum number of tweet by an individual was 570. We show the histogram for only up-to
% ten tweets due to space constraints.
% As may be expected, the histogram follows the power law. 
% We chose to annotate only one tweet per tweeter to avoid individual biases.\\ \\ \\ \\

% \subsection{Analysis of the 2012 US Election Tweets Corpus}
% \hspace{3mm} -- distribution as per hashtag

% \begin{table}[t]
% \caption{\label{tab:histo} A portion of the histogram of the number of tweets per person.}
% \begin{center}
% {\small
%  \begin{tabular}{lr}
% \hline \bf Number of tweets & \bf Number of \\ 
% \bf per person & \bf people \\ \hline
% 1   &61866\\
% 2   &8027\\
% 3   &3020\\
% 4   &1541\\
% 5   &971\\
% 6   &667\\
% 7   &446\\
% 8   &351\\
% 9   &273\\
% 10  &262\\
% \hline
% \end{tabular}
% }
% \end{center}
% \end{table}
%

\subsection{Annotating Purpose by Crowdsourcing}
\label{sec:annotation}

We used Amazon's Mechanical Turk service to crowdsource the annotation of the electoral tweets.\footnote{https://www.mturk.com/mturk/welcome}
We randomly selected about 2,000 tweets, each  by a different Twitter user. 
% Further, we selected tweets from only those people who contributed exactly one tweet in our collection. 
We asked  a series of questions for each tweet.
Below is the questionnaire for an example tweet:

{\small
{\bf Purpose behind US election tweets}

%  {\bf Instructions}
%  \begin{itemize}
%  \item Attempt HITs only if you are a native speaker of English. %, or very fluent in English.
%  \item Certain check questions will be used to make sure the annotation is responsible and reasonable. Assignments that fail these tests will be rejected.
%  \item Your responses are confidential.  Any publications based on these responses will not include your specific responses, but rather, aggregate information from many individuals.% We will not ask any information that can be used to identify who you are.
%  \end{itemize}
\vspace*{-2mm}
 {\bf Tweet:} Mitt Romney is arrogant as hell.

\vspace*{-2mm}
 Q1. Which of the following best describes the purpose of this tweet? 
% \begin{itemize}
% \item to point out hypocrisy or inconsistency
% \item to point out mistake or blunder
% \item to disagree
% \item to ridicule
% \item to criticize, but none of the above 
% \item to vent\\
%  ---
% \item to agree
% \item to praise, admire, or appreciate 
% \item to support
% \item to motivate or to incite action\\
% % \item to be entertaining [I think we want to delete all mentions of this option. Leaving it here for now.]
% ---
% \item to provide information without emotional content
% \item none of the above
% \end{itemize}
\vspace*{-4mm}
\begin{quote}
\hspace*{-0mm} - to point out hypocrisy or inconsistency\\
\hspace*{-0mm} - to point out mistake or blunder\\
\hspace*{-0mm} - to disagree\\
\hspace*{-0mm} - to ridicule\\
\hspace*{-0mm} - to criticize, but none of the above \\
\hspace*{-0mm} - to vent
\end{quote}
\vspace*{-6mm}
\begin{quote}
\hspace*{-0mm} - to agree\\
\hspace*{-0mm} - to praise, admire, or appreciate\\ 
\hspace*{-0mm} - to support
%\hspace*{-4mm} - to motivate or to incite action
\end{quote}
\vspace*{-6mm}
\begin{quote}
\hspace*{-0mm} - to provide information without emotion\\
\hspace*{-0mm} - none of the above
\end{quote}

\vspace*{-4mm}
Q2. Is this tweet about US politics and elections? 
\vspace*{-4mm}
\begin{itemize}
\item Yes, this tweet is about US politics and elections.
\item No, this tweet has nothing to do with US politics or anybody involved in it.
\end{itemize}
}
\vspace*{-4mm}
\noindent These questionnaires are called {\it HITs (human intelligence tasks)} in Mechanical Turk parlance. 
We posted 2042 HITs corresponding to 2042 tweets. We requested responses from at least three annotators
for each HIT. % 2096
The response to a HIT by an annotator is called an {\it assignment}.
In Mechanical Turk, an annotator may provide assignments for as many HITs as they wish. 
Thus, even though only three annotations are requested per HIT, 
about 400 annotators contribute assignments for 
the 2,042 tweets. The number of assignments completed by the annotators followed a zipfian distribution.

Even though it is possible that more than one option may apply for a tweet, we allowed the Turkers
to select only one option for each question. We did this to encourage annotators to select the
option that best answers the questions. We wanted to avoid situations where an annotator selects multiple
options just because they are vaguely relevant to the question.

\begin{table}[t]
\caption{\label{tab:tweetdist} The histogram of the number of annotations of tweets. `annotns' is short for annotations.}
\begin{center}
\begin{tabular}{crr}
%\hline \bf \# of annotns.\@ & & \\ 
\hline \bf annotns/tweet & \bf \# of tweets & \bf \# of annotns\\ \hline
1 &181	&181\\
2 &594	&1188\\
3	&1121	&3363\\
4	&60		&240\\	
$\ge$5	&88	&1509\\% \hline
all & 2042 &6481\\
\hline
\end{tabular}
\end{center}
\vspace*{-6mm}
\end{table}

Since there has been no prior study on classifying electoral tweets into different classes of purpose,
we asked our colleagues for the main kinds of purpose by providing them with some example election tweets.
The final list of eleven categories was selected from their responses.
Observe that we implicitly grouped the options for Q1 into three coarse categories 
by putting extra vertical space between the groups. These coarse categories correspond to {\it oppose} 
(to point out hypocrisy, to point out mistake, to disagree, to ridicule, to criticize, to vent),
{\it favour} (to agree, to praise, to support), and {\it other}.
Even though there is some redundancy among the fine categories,
they are more precise and may help annotation. Eventually, however,
it may be beneficial to combine two or more categories for the purposes of automatic classification. 
The amount of combining will depend on
the task at hand, and can be done to the extent that anywhere from eleven to two categories remain.
% it may be sufficient to obtain information about the coarser categories.

\subsection{Annotation Analyses}
\label{sec:ann_analysis}

The Mechanical Turk annotations were done over a period of one week.
For each annotator, and for each question, we calculated the probability with which the annotator agrees
with the response chosen by the majority of the annotators.
We identified poor annotators as those that had an agreement probability that was
more than two standard deviations away from the mean. All annotations by these
annotators were discarded. 
% All analysis below pertains to the remaining data,
% which comprises of two or more independent annotations for 2042 tweets.
Table \ref{tab:tweetdist} gives a histogram of the number of annotations of the remaining tweets.
There were 1121 tweets with exactly three annotations.

% \begin{table}[t]
% \begin{center}
% \begin{tabular}{rrr}
% \hline \bf \# of annotns. & \bf \# of tweets & \bf \# of annotns.\\ \hline
% 1 &22	&22\\
% 2 &458	&916\\
% 3	&1450	&2900\\
% 4	&78		&312\\	
% $\ge$5	&88	&3066\\\hline
% all & 2096 &7216\\
% \hline
% \end{tabular}
% \end{center}
% \caption{\label{tab:tweetdist} The histogram of the number of annotations of tweets. `annotns.' stands for annotations.}
% \end{table}

%% majority vote
\begin{table}[t]
\caption{\label{tab:distpurposemaj} Percentage of tweets in each of the eleven categories of Q1. 
Only those tweets that were annotated by at least two annotators were included.  A tweet belongs to category X if it is annotated with X more often than all other categories combined.  There were 1072 such tweets in total.}
\begin{center}
\resizebox{0.47\textwidth}{!}{
\begin{tabular}{lr}
\hline  & \bf Percentage\\ 
 \bf Purpose of tweet & \bf of tweets \\ \hline
favour & \\
\ \ to agree & 0.47\\
\ \ to praise, admire, or appreciate & 15.02 \\
\ \ to support & {\bf 26.49}\\
% & \\
oppose & \\
\ \ to point out hypocrisy or inconsistency & 7.00 \\
\ \ to point out mistake or blunder & 3.45 \\
\ \ to disagree & 2.52 \\
\ \ to ridicule & 15.39\\
\ \ to criticize, but none of the above &7.09 \\
\ \ to vent & 8.21 \\
% & \\
other & \\
\ \ to provide information without any  & \\
\ \ \ \ emotional content 				& 13.34 \\
\ \ none of the above & 1.03 \\%\hline
% Total & 1146 \\
	all & 100.0 \\
\hline
\end{tabular}
}
\end{center}
% A tweet belongs to category X if it was annotated with X more often than with all other categories altogether. 
% Category ``to support'' includes ``to motivate or to incite action''.}
\vspace*{-6mm}
\end{table}

\begin{table}[t]
\caption{\label{tab:3class-dist} Percentage of tweets in each of the three coarse categories of Q1.
 Only those tweets that were annotated by at least two annotators were included.
 A tweet belongs to category X if it is annotated with X more often than all other categories combined.
 There were 1672 such tweets in total.
%  The total number of tweets is larger since the annotator 
Agreement on the 3 categories is higher than on 11 categories.}
\begin{center}
\begin{tabular}{lr}
\hline  % & \bf Percentage\\
\bf Category & \bf Percentage of tweets \\ \hline
oppose & \bf 58.07 \\
favour 	& 31.76 \\
other 		& 10.17 \\ %\hline
all & 100.0 \\
\hline
\end{tabular}
\end{center}
\vspace*{-6mm}
\end{table}

% Tables \ref{tab:distpurpose} and \ref{tab:distpol} give the distributions of the various options for questions 1 and 2, respectively.
% % These numbers are calculated from the total set of annotations for the 2042 tweets.
% For example, tables \ref{tab:distpurpose} shows that 19.7\% of the annotations identified the target tweets to have the purpose `to support'.

We determined whether a tweet is to be assigned a particular category based on strong majority.
That is, a tweet belongs to category X if it is annotated with X more often than all other categories combined.
% Only those tweet that were annotated by at least two annotators were included. 
Percentage of tweets in each of the 11 categories of Q1 are shown in Table \ref{tab:distpurposemaj}.
Observe that the majority category for purpose is `to support'---26.49\% of the tweets were identified as having the purpose `to support'.
Table \ref{tab:3class-dist} gives the distributions of the three coarse categories of purpose.
Observe, that the political tweets express disagreement (58.07\%) much more than support (31.76\%).

Table \ref{tab:distpolmaj} gives the distributions for question 2.
Observe that a large majority (95.56\%) of the tweets are relevant to US politics and elections.
This shows that the hashtags shown earlier in Table \ref{tab:poltags} are effective in identifying political tweets.

\begin{table}[t]
\caption{\label{tab:distpolmaj} Percentage of tweets in each of the two categories of Q2.}
\begin{center}
\resizebox{0.47\textwidth}{!}{
\begin{tabular}{lr}
\hline  & \bf Percentage\\ 
 \bf Relevance & \bf of tweets \\ \hline
	pertaining to US politics and elections 						&{\bf 95.56}\\
	not pertaining to US politics and elections	 	&4.44\\%\hline
	all & 100.0 \\
\hline
\end{tabular}
}
\end{center}
% Only those tweets that were annotated by at least two annotators were included.}
% A tweet belongs to category X if it is annotated with X more often than all other categories combined.
% There were 1829 such tweets in total.}
\end{table}

\subsubsection{Inter-Annotator Agreement}
\label{sec:IAA}
We calculated agreement on the full set of annotations, and not just
on the annotations with a strong majority as described in the previous section.
 One way to gauge the amount of agreement among annotators is to
 examine the number of times all three annotators agree (majority class
 size = 3), the number of times two out of three annotators agree
 (majority class size = 2), and the number of times all three
 annotators choose different options (majority class size = 1).

 Table \ref{tab:mcs} gives the distributions of the majority classes. Higher numbers for the larger class
 sizes indicate higher agreement.
 For example, for 22.4\% of the tweets all three annotators gave the same answer for question 1 (Q1).
 The agreement is much higher if one only considers the coarse categories of `oppose', `favour', and `other'---these
 numbers are shown in the row marked Q1'.
 The agreement for question 2 was substantially high. This was expected as it is a relatively straightforward question.
 The numbers in the table are calculated from tweets with exactly three annotations.

% \begin{table}[t]
% \begin{center}
% \begin{tabular}{lrrr}
% \hline  & \bf MCS-1 & \bf MCS-2 & \bf MCS-3 \\ \hline
% q1      &4.0        &30.0       &66.0\\
% q2      &11.6       &42.7       &45.7\\
% q3      &29.5       &48.1       &22.4\\
% q4      &0.0        &5.7        &94.3\\
% q5      &2.2        &31.7       &66.1\\
% \hline
% \end{tabular}
% \end{center}
% \caption{\label{tab:mcs} Percentage of tweets having majority class size (MCS) 1, 2, and 3.
% q stands for question.}
% \end{table}

 \begin{table}[t]
 \caption{\label{tab:mcs} Percentage of tweets having majority class size (MCS) of 1, 2, and 3.  Q is short for question.}
 \begin{center}
 \begin{tabular}{lrrr}
 \hline  & \bf MCS-1 & \bf MCS-2 & \bf MCS-3 \\ \hline
 Q1      &29.5       &48.1       &22.4\\
 Q1'     &2.2        &31.7       &66.1\\
 Q2      &0.0        &5.7        &94.3\\
 \hline
 \end{tabular}
 \end{center}
 \end{table}

Table \ref{tab:ia} shows {\it inter-annotator agreement (IAA)}, for the two questions---the average percentage of times two annotators
agree with each other.
IAA gives us an understanding of the degree of agreement through a single number.
Observe that the agreement is only moderate for the eleven fine categories of purpose (43.58\%), but much higher
when considering the coarser categories (83.81\%).
% [Need to include Scott's Pie scores.]

% \begin{table}[t]
% \begin{center}
% \begin{tabular}{lrrr}
% \hline
%   &IAA\\
% \hline
% q1      &78.02\\
% q2      &55.77\\
% q3      &43.58\\
% q4      &96.76\\
% q5      &83.81\\
% \hline
% \end{tabular}
% \end{center}
% \caption{\label{tab:ia} Inter-annotator agreement.}
% \end{table}

\begin{table}[t]
\caption{\label{tab:ia} Agreement statistics: inter-annotator agreement (IAA) and average probability of choosing the majority class (APMS).}
\begin{center}
\begin{tabular}{lrrr}
\hline
  &\bf IAA	&\bf APMS\\
\hline
% q1      &78.02\\
% q2      &55.77\\
Q1      &43.58		&0.520\\
Q1'     &83.81		&0.855\\
Q2      &96.76		&0.974\\
\hline
\end{tabular}
\end{center}
% \vspace*{-3mm}
\end{table}

Another way to gauge agreement is by calculating the average
probability with which an annotator picks the majority class.
Consider the example below:
Each tweet is annotated by 3 different annotators.
X annotates 10 tweets.
Six of the times, X's answer for Q1 is the answer that has a majority
(in case of 3 annotators, this means that at least one other annotator
also gave the same answer as X for 6 of the 10 tweets). Thus the probability
with which X picks the majority class is 6/10.
The last column in Table \ref{tab:ia} shows the {\it average probability of picking the majority class (APMS)}
by the annotators
% by taking the sum of the numerators for all of
% the annotators and dividing by the sum of the denominators. This micro
% average makes sense for all tweets with more than one annotation, and
% it gives us a sense of inter-annotator agreement through a single
% number for each question.
% Below are the probabilities 
(higher numbers indicate higher agreement).

% \begin{table}[t]
% \begin{center}
% \begin{tabular}{lr}
% \hline
%   &APMS\\
% \hline
% q1      &0.845\\
% q2      &0.688\\
% q3      &0.520\\
% q4      &0.974\\
% q5      &0.855\\
% \hline
% \end{tabular}
% \end{center}
% \caption{\label{tab:apms} Average probability of choosing the majority class (APMS).}
% \end{table}

% \begin{table}[t]
% \begin{center}
% \begin{tabular}{lr}
% \hline
%   &APMS\\
% \hline
% Q1      &0.520\\
% Q1'     &0.855\\
% Q2      &0.974\\
% \hline
% \end{tabular}
% \end{center}
% \caption{\label{tab:apms} Average probability of choosing the majority class (APMS).}
% \end{table}

Overall, we observe that there is strong agreement between annotators at identifying whether
the purpose of a tweet is to oppose, to favour, or something else.

 \begin{table*}[t]
 \caption{\label{tab:cm-q1} Confusion Matrix: Question 1 (fine-grained). The value in a particular cell, say for row x and column y, is the number of annotations
  that were assigned label y even though the majority votes for each of those tweets were for x. The highest number in each row is shown in bold.}
 \begin{center}
 {\small
 \begin{tabular}{lr rrr rrr rrr rr}
 \hline
                         & & \bf c1     &\bf c2     &\bf c3     &\bf c4     &\bf c5     &\bf c6     &\bf c7     &\bf c8     &\bf c9     &\bf c10    &\bf c11    \\
\hline
 favour &\\
 \ \ to agree: &r1    &{\bf 20}     &5      &9      &2      &1      &2      &0      &3      &0      &4      &0      \\
 \ \ to praise, admire, or appreciate: &r2    &0      &{\bf 291}    &61     &1      &1      &5      &1      &5      &4      &3      &0      \\
 \ \ to support: &r3  &1      &43     &{\bf 565}    &5      &4      &23     &7      &18     &5      &22     &3      \\
 & & & &\\
 oppose &\\
 \ \ to point out hypocrisy or inconsistency: &r4     &2      &2      &14     &{\bf 123}    &15     &26     &10     &64     &11     &5      &0      \\
 \ \ to point out mistake or blunder: &r5     &0      &6      &16     &6      &{\bf 84}     &29     &15     &46     &1      &3      &0      \\
\ \ to disagree: &r6         &0      &0      &5      &10     &2      &{\bf 145}    &10     &5      &5      &1      &0      \\
 \ \ to ridicule: &r7         &3      &11     &28     &9      &16     &37     &{\bf 274}    &60     &15     &4      &0      \\
 \ \ to criticize, but none of the above: &r8         &1      &0      &22     &8      &5      &49     &30     &{\bf 227}    &9      &3      &0      \\
 \ \ to vent: &r9     &7      &12     &35     &5      &11     &37     &22     &45     &{\bf 155}    &7      &1      \\
 & & & &\\
 other &\\
 \ \ to provide information without any       & & & & & & & & & & &\\
 \ \ \ \ \ emotional content: &r10       &2      &11     &39     &1      &4      &8      &11     &19     &8      &{\bf 259}    &4      \\
 \ \ none of the above: &r11  &3      &6      &10     &1      &4      &5      &7      &3      &6      &10     &{\bf 19}     \\
 \hline
 \end{tabular}
 }
 \end{center}
 \end{table*}

 \subsubsection{Confusion Matrix}
 
 Human annotators may disagree with each other because 
 %of error in judgment on the part of one or more annotators or because 
 two or more options may seem appropriate for a given tweet. 
There also exist tweets where the purpose is unclear. 
 Table \ref{tab:cm-q1} shows the confusion matrix for question 1. The rows and columns
 of the matrix correspond to the eleven options. The value in a particular cell, say for row x and column y, is the number of annotations
 that were assigned label y even though the majority votes for each of those tweets were for x.
The highest number in each row is shown in bold.
 The cells in the diagonal correspond to the number of instances for which the annotations matched the majority vote.
 For high agreement, one would want higher numbers in the diagonal, which is what we observe in Table \ref{tab:cm-q1}.

 We can identify options that tend to be confused for each other by noting non-diagonal cells with high values.
 For example, consider cell r7--c8. The relatively large number indicates that `to ridicule' is often confused
  with `to criticize, but none of the above'. Similarly, we find that `to point out hypocrisy or inconsistency'
 and `to point out mistake or blunder' are also often confused with `to criticize, but none of the above' (r4--c8 and r5--c8).
This suggests that the purpose of `to criticize, but none of the above' is relatively harder to identify.
Note however, that the labels are not confused as strongly in the other direction. That is, tweets that
have a purpose of `to criticize' are not confused as much with `to point out hypocrisy or inconsistency' (r8--c4), `to point out mistake or blunder' (r8--c5), or `to ridicule' (r8--c7).
Thus there is some clear signal even in `to criticize, but none of the above' that human annotators are able to exploit.
 Among the categories of favor, `to praise, admire, or appreciate' is confused with `to support'.
 This suggests that the category `to criticize, but none of the above' serves as a
 hold-back for other finer-grained categories of `oppose' and, therefore, is often chosen
 by annotators for less clear messages. A similar situation occurs in the `favour' group,
 where the confusion occurs mostly between a more general category `to support' and more
 specific categories `to agree' and `to praise, admire, or appreciate'.  Note that in a
 particular application, one may choose only a subset of the eleven categories that are
 most relevant.  For example, one may combine `to point out hypocrisy or inconsistency',
 `to point out mistake or blunder', and `to criticize, but none of the above' into a
 single category, and distinguish it from other oppose categories such as `to disagree'
 and `to ridicule'.  Table \ref{tab:cm-q1coarse} shows the confusion matrix within the
 coarse categories of question 1.  The confusion between the coarse categories is
 relatively lower than among the finer categories, but yet there exist instances when
 `favour' is confused with `oppose', and vice versa.  Table \ref{tab:cm-q2} shows the
 confusion matrix for question 2. Only a very small number of instances are confused with
 the wrong option for this question.

 \begin{table}[t]
 \caption{\label{tab:cm-q1coarse} Confusion Matrix: Question 1' (coarse grained).}
 \begin{center}
 {\small
 \begin{tabular}{lr rrr}
 \hline
 & 	&\bf c1     &\bf c2     &\bf c3     \\
 \hline
 
 favour: &r1   &{\bf 941}    &136    &37     \\
 oppose: &r2   &75     &{\bf 1705}   &29     \\
 other: &r3   &40     &88     &{\bf 312}    \\
 \hline
 \end{tabular}
 }
 \end{center}
 \end{table}

 \begin{table}[t]
 \caption{\label{tab:cm-q2} Confusion Matrix: Question 2.}
 \begin{center}
 \resizebox{0.48\textwidth}{!}{
 % {\small
 \begin{tabular}{lr rr}
 \hline
       &                  &\bf c1     &\bf c2     \\
 \hline
 not pertaining to US politics and elections: &r1        &{\bf 106}    &38     \\
 pertaining to US politics and elections: &r2         &26     &{\bf 3193}   \\
 \hline
 \end{tabular}
 }
 \end{center}
 \end{table}

\begin{table*}[t]
\caption{\label{tab:purpose-emotion} Percentage of different purpose tweets pertaining to different emotions. 
Low-frequency categories of purpose and emotion are omitted. The highest number for each category of purpose is shown in bold.}
\begin{center}
\resizebox{\textwidth}{!}{
% {\small
\begin{tabular}{lrrrrrrr}
\hline & \bf admiration & \bf	anticipation & \bf	joy & \bf dislike & \bf	disappointment & \bf	disgust & \bf	anger \\ \hline
favour & \\
\ \ to praise, admire, or appreciate	& \bf	67	& 4	& 25 & & & & \\
\ \ to support	& \bf	33	& 21	& 21	& 4 &	2	& &	7 \\
& & &\\
oppose & \\
\ \ to point out hypocrisy or inconsistency	& & & &	\bf	61 &	& 17 &	11\\
\ \ to point out mistake or blunder	& & & &	\bf	77 &	& 15	& 8 \\
\ \ to disagree	& & &	14	& \bf	43 & &	14 &	29 \\
\ \ to ridicule	& & &	7	& \bf	66	& &	7	& 18 \\
\ \ to criticize, but none of the above	& & &	& \bf	47	& 11	& 16	& 16 \\
\ \ to vent	& & &	4 &	24 &	12	& 8	& \bf	36\\ \hline
\end{tabular}
}
\end{center}
\end{table*}

\subsection{Distinctions between purpose and affect}

The task of detecting purpose is related to sentiment and emotion
classification. Intuitively, the three broad categories of purpose, `oppose', `favour', and `other', roughly correspond to negative, positive,
and objective sentiment. Also, some fine-grained categories seem to partially
correlate with emotions. For example, when angry, a person vents. When overcome
with admiration, a person praises the object of admiration. 
In our experiments, we showed that resources  created for emotion detection helped
identify purpose.

To further investigate the relation between purpose and emotion, we 
annotated a portion of the tweets 
by crowdsourcing with one of 19 emotions:
acceptance, admiration, amazement, anger,
anticipation, calmness, disappointment, disgust, dislike, fear, hate,
indifference, joy, like, sadness, surprise, trust, uncertainty, and vigilance.
Similar to the annotation of purpose, each tweet was annotated by
at least two judges, and tweets with no strong majority were discarded. 

Table~\ref{tab:purpose-emotion} shows the percentage of tweets pertaining to
different emotions. Only high-frequency categories of
purpose and emotion are shown. As expected, the tweets with the purpose 
`favour' mainly convey the emotions of admiration, anticipation, and
joy. On the other hand, the tweets with the purpose `oppose' are mostly
associated with negative emotions such as dislike, anger, and disgust.
The purpose `to praise, admire, or appreciate' is highly correlated with
the emotion admiration.
% whereas `to support' is quite evenly split among `admiration', `anticipation', and `joy'. 

Note that most of the tweets with the purpose `to point out hypocrisy', `to point out mistake',
`to disagree', `to ridicule', `to criticize', and even many instances of `to vent'
are associated with the emotion dislike.
Thus, a system that only determines emotion and not purpose will fail to distinguish
between these different categories of purpose.
% , with the exception of category `to vent' in which the majority of the tweets express `anger'.
It is possible for people to have the same emotion of dislike and react differently:
either by just disagreeing, pointing out the mistake, criticizing, or resorting to ridicule.

\section{Automatically Identifying Purpose}
To automatically classify tweets into eleven categories of purpose, we trained a
Support Vector Machine (SVM) classifier. SVM is a state-of-the-art learning
algorithm proved to be effective on text categorization tasks and robust on
large feature spaces. The eleven categories were assumed to be mutually
exclusive, i.e., each tweet was classified into exactly one category. In the
second set of experiments, the eleven fine-grained categories were combined into
3 coarse-grained - `oppose', `favour', and `other' - as was
described earlier. In each experiment, ten-fold stratified cross-validation was
repeated ten times, and the results were averaged. 
Paired t-test was used to confirm the significance of the results.
We used the LibSVM
package \cite{CC01a} with linear kernel and default parameter settings. Parameter
C was chosen by cross-validation on the training portion of the data (i.e., the
nine training folds). 
%Accuracy was used as the evaluation measure:
%	\[
%Accuracy = \frac{\text{number of correctly predicted instances}}{\text{total number of instances}}
%\]
%Note that in the case of mutual exclusive categories, accuracy equals to micro-averaged precision, recall, and F1-measure.
 
The gold labels were determined by strong majority voting.
Tweets with less than 2 annotations or with no majority labels were discarded. Thus, the
dataset consisted of 1072 tweets for the 11-category task, and 1672 tweets for
the 3-category task. The tweets were normalized by replacing all URLs with
http://someurl and all userids with @someuser. The tweets were tokenized
and tagged with parts of speech using the Carnegie Mellon University 
Twitter NLP tool \cite{Gimpel11}.

\subsection{A Basic System for Purpose Classification}
 
Each tweet was represented as a feature vector with the following groups of features.
We drew these features from prior work on social media and sentiment analysis \cite{PangL08,Barbosa10,Rao10}. %,Agarwal11}
We employed commonly used text classification features such as ngrams, part-of-speech, and punctuations, as well as common
Twitter-specific features such as emoticons and hashtags. Additionally, we hypothesized that 
the purpose of tweets is guided by the emotions of the tweeter. Thus we explored certain emotion features as well.

	\vspace*{-4mm}
\begin{itemize}
	\item n-grams: presence of n-grams (contiguous sequences of 1, 2, 3, and 4 tokens), skipped n-grams (n-grams with one token replaced by *), character n-grams (contiguous sequences of 3, 4, and 5 characters);
	\vspace*{-2mm}
	\item POS: number of occurrences of each part-of-speech;
	\vspace*{-2mm}
	\item word clusters: presence of words from each of the 1000 word clusters 
provided by the Twitter NLP tool \cite{Gimpel11}.
These clusters were	produced with the Brown clustering algorithm on 56 million English-language tweets. 
They serve as alternative
representation of tweet content, reducing the sparcity of the token space;
	\vspace*{-2mm}
	\item all-caps: the number of words with all characters in upper case;
	\vspace*{-2mm}
	\item NRC Emotion Lexicon: 	We used the NRC Emotion Lexicon \cite{MohammadT10} to incorporate affect features.
The lexicon consists of 14,182 words manually annotated with 8 basic
emotions (anger, anticipation, disgust, fear, joy, sadness, surprise, trust) and
2 polarities (positive, negative). Each word can have zero, one, or more
associated emotions and zero or one polarity.
	\vspace*{-2mm}
        \begin{itemize}
	         \item number of words associated with each emotion
	\vspace*{-1mm}
	         \item number of nouns, verbs, etc., associated with  each emotion
	\vspace*{-1mm}
	         \item number of all-caps words associated with each emotion
	\vspace*{-1mm}
	         \item number of hashtags associated with each emotion
        \end{itemize}
	\vspace*{-2mm}
	\item negation: the number of negated contexts. Following \cite{PangLV02}, we defined a negated context as a segment of a tweet that starts with a negation word (e.g., `no', `shouldn't') and ends with one of the  
	punctuation marks: `,', `.', `:', `;', `!', `?'. A negated context affects the n-gram and Emotion Lexicon features: each word and associated with it emotion in a negated context become negated 
	(e.g., `not perfect' becomes `not perfect\_NEG', `EMOTION\_trust' becomes `EMOTION\_trust\_NEG').
The list of negation words was adopted from Christopher Potts' sentiment tutorial.\footnote{http://sentiment.christopherpotts.net/lingstruc.html}
	\vspace*{-2mm}
	\item punctuation: the number of contiguous sequences of exclamation marks, question marks, and both exclamation and question marks;
	\vspace*{-2mm}
	\item emoticons: presence/absence of positive and negative emoticons. The polarity of an emoticon was determined with a simple regular expression
	adopted from Christopher Potts' tokenizing script.\footnote{http://sentiment.christopherpotts.net/tokenizing.html}
	\vspace*{-2mm}
	\item hashtags: the number of hashtags;
	\vspace*{-2mm}
	\item elongated words: the number of words with one character repeated more than 2 times, e.g.  `soooo'.
\end{itemize}

% \subsection{Experiments}

\begin{table}[t]
\caption{\label{tab:results-bounds} Accuracy of the automatic classification on 11-category and 3-category problems. The lower bound is the percentage of the majority class. }
\begin{center}
\begin{tabular}{lll}
\hline & \bf 11-class & \bf 3-class \\ \hline
majority class & 26.49 & 58.07 \\
SVM & 43.56 & 73.91 \\
% human annotators & 79.43 & 91.57\\
\hline
\end{tabular}
\end{center}
\end{table}

Table~\ref{tab:results-bounds} presents the results of the automatic
classification for the 11-category and 3-category problems. For comparison, we
also provide the 
accuracy of a simple baseline classifier that always predicts the majority
class. 
% The upper bound is the accuracy of human annotations
% calculated on the same dataset. We can see that automatic classifier outperforms
% the baseline classifier by a large margin, yet it does not match human
% performance. Note, however, that the upper bound is probably overly optimistic
% as the same human annotations formed the gold standard through the majority
% voting. 

\begin{table}[t]
\caption{\label{tab:results-per-class} Per category precision (P), recall (R), and F1 score of the classification on the 11-category problem. 
Micro-averaged P, R, and F1 are equal to accuracy since the categories are mutually exclusive.}
\begin{center}
\resizebox{0.47\textwidth}{!}{
% {\small
\begin{tabular}{lrrrr}
\hline \bf category & \bf \# inst. & \bf P & \bf R & \bf F1 \\ \hline
favour & \\
\ \ to agree	& 5 & 0 &	0 & 	0\\
\ \ to praise	& 161 & 57.59	& 50.43	& 53.77\\
\ \ to support	& 284 & 49.35 &	69.47	& 57.71\\
% & & & &\\
oppose & \\
\ \ to point out hypocrisy	& 75 & 30.81	& 21.2	& 25.12\\
\ \ to point out mistake	& 37 & 0	& 0	 & 0\\
\ \ to disagree	& 27 & 0	& 0	 & 0\\
\ \ to ridicule	& 165 & 31.56	& 43.76	& 36.67\\
\ \ to criticize	& 76 & 22.87	& 9.87	& 13.79\\
\ \ to vent	& 88 & 36.06	& 23.07	& 28.14\\
% & & & &\\
other & \\
\ \ to provide information	& 143 & 45.14	& 50.63	& 47.73\\
\ \ none of the above	& 11 & 0	& 0	& 0\\ % \hline
 micro-ave & & 43.56 & 43.56 & 43.56\\
\hline
\end{tabular}
}
\end{center}
\end{table}

Table~\ref{tab:results-per-class} shows the classification results broken-down
by category. As expected, the categories with larger amounts of labeled examples
(`to praise', `to support', `to provide information') have higher results. 
However, for one of the higher frequency categories, `to ridicule’, the F1-score is relatively low. This category incorporates irony, sarcasm, and humour, the concepts that are hard to recognize, especially in a very restricted context of 140 characters.
The
four low-frequency categories (`to agree', `to point out mistake or blunder',
`to disagree', `none of the above') did not have enough training data for the
classifier to build adequate models. The categories within
`oppose' are more difficult to distinguish among than the
categories within `favour'. However, for the most part this can be
explained by the larger number of categories (6 in `oppose' vs. 3 in `favour') and, consequently, smaller sizes of the individual categories.

In the next set of experiments, we investigated the usefulness of each feature
group for the task. We repeated the above classification process, each time
removing one of the feature groups from the tweet representation.
Table~\ref{tab:ablation-results} shows the results of these ablation experiments
for the 11-category and 3-category problems. In both cases, the most influential
features were found to be n-grams, emotion lexicon features, part-of-speech
tags, and word clusters.

\begin{table}[t]
\caption{\label{tab:ablation-results} Accuracy of classification with one of the feature groups removed. Numbers in bold represent statistically significant difference with the accuracy of the `all features' classifier (first line) with 95\% confidence.}
\begin{center}
\resizebox{0.47\textwidth}{!}{
\begin{tabular}{lll}
\hline \bf Experiment & \bf 11-class & \bf 3-class \\ \hline
all features  &  43.56 & 73.91 \\
all - n-grams & \bf 39.51 & \bf 71.02 \\
all - NRC emotion lexicon & \bf 42.27 & \bf 72.21\\
all - parts of speech & \bf 42.63 & \bf 73.55\\
all - word clusters & \bf 43.24 & \bf 73.24\\
all - negation & \bf 43.18 & \bf 73.36 \\
all - (all-caps, punctuation,  & \\
emoticons, hashtags) & 43.38 & 73.87\\
% emoticons, hashtags, elongated words) & 43.38 & 73.87\\
\hline
\end{tabular}
}
\end{center}
\vspace*{-2mm}
\end{table}

\subsection{Adding features pertaining to hundreds of fine emotions}

\begin{table}[t]
\caption{\label{tab:lexicon-results} Accuracy of classification using different lexicons on the 11-class problem. Numbers in bold represent statistically significant difference with the accuracy of the classifier using the NRC Emotion Lexicon (first line) with 95\% confidence.}
\begin{center}
\begin{tabular}{lr}
\hline \bf Lexicon & \bf Accuracy \\ \hline
NRC Emotion Lexicon & 43.56 \\
Hashtag Lexicon & 44.35 \\
both lexicons & \bf 44.58 \\ \hline
\end{tabular}
\end{center}
\vspace*{-4mm}
\end{table}

Since the emotion lexicon had a significant impact on the results, we further created 
% a a bigger lexical
% resource would bring additional advantage to automatic identification of
% purpose. Thus, we constructed 
a wide-coverage twitter-specific lexical resource following on work by Mohammad \shortcite{Mohammad12}. 
\cite{Mohammad12} showed that emotion-word hashtagged tweets are a good source of labeled data
for automatic emotion processing. Those experiments were conducted using tweets pertaining
to the six Ekman emotions because labeled evaluation data exists for only those emotions.
However, a significant advantage of using hashtagged tweets is that we can collect
large amounts of labeled data for any emotion that is used as a hashtag by tweeters.
Thus we polled the Twitter API and collected a large corpus of tweets pertaining
% not only to the Robert Plutchik set of eight basic emotions (which subsumes Ekman's six),
% but also tweets pertaining 
to a few hundred emotions.

We used a list of 585 emotion words compiled by Zeno G. Swijtink as the hashtagged query words.\footnote{http://www.sonoma.edu/users/s/swijtink/teaching/\\philosophy\_101/paper1/listemotions.htm}
Note that we chose not to dwell on the question of whether each of the words in this set is truly an emotion
or not. Our goal was to create and distribute a large set of affect-labeled data, and users are
free to choose a subset of the data that is relevant to their application.
% A list of 585 emotion-related hashtags (e.g., \#love, \#annoyed,
% \#pity) was compiled from different sources. Then, a large number of tweets
% containing at least one of these hashtags were collected from Twitter. 
We calculated the
pointwise mutual information (PMI) between an emotional hashtag and a word
appearing in tweets. The PMI represents a degree of correlation between the word and
emotion, with larger scores representing stronger correlations. Consequently,
the pairs (word, hashtag) that had positive PMI were pulled together into a new
word--emotion association resource, that we call {\it Hashtag Emotion Lexicon}. The lexicon
contains around 10,000 words with associations to 585 emotion-word hashtags. 

We used the Hashtag Lexicon for classification by creating a separate feature for each
emotion-related hashtag,  resulting in 585 emotion features. The values of these
features were calculated as the sum of the PMI scores between the words in a
tweet and the corresponding emotion-related hashtag.
Table~\ref{tab:lexicon-results} shows the results of the automatic
classification using the new lexical resource. The Hashtag Lexicon 
significantly improved the performance of the classifier on the 11-category task. 
Even better
results were obtained when both lexicons were employed\footnote{Using the Hashtag Lexicon pertaining to hundreds of emotions on the 3-category task did not show any improvement. 
This is probably because there the information about positive and negative sentiment provides the most gain.}.

\section{Conclusions}
Tweets are playing a growing role in the public discourse on politics.
In this paper, we explored the purpose behind such tweets.
 Detecting purpose has a number of applications including detecting the mood of the electorate, estimating the popularity of policies,
 identifying key issues of contention, and predicting the course of events.
We compiled a dataset of 1 million tweets pertaining to the 2012 US presidential elections using relevant hashtags.
We designed an online questionnaire and annotated a few thousand tweets
for purpose via crowdsourcing.
We analyzed these tweets and showed that a large majority convey emotional attitude towards someone or something.
Further, the number of messages posted to oppose someone or something were almost twice the number of
messages posted to offer support.

We developed a classifier to automatically classify electoral tweets as per their purpose.
It obtained an accuracy of 43.56\% on a 11-class task and an accuracy of 73.91\% on a 3-class task (both accuracies
well above the most-frequent-class baseline).
% We show that identifying purpose is correlated with the related phenomenon of sentiment and emotion detection, but yet significantly different. 
We found that resources developed for emotion detection, such as the NRC word--emotion association lexicon,
are also helpful for detecting purpose. 
However, we also showed that emotion detection alone can fail to distinguish between several kinds of purpose.
% In the process, we create a new word--emotion association lexicon with information about hundreds of emotions
% that has the potential for use in various affect-analysis tasks.
We make all the data created as part of this research freely available.

In this paper, we relied only on the target tweet as context. However, it might be possible to obtain even better
results by modeling user behaviour based on multiple past tweets. 
% Another avenue for future research
% is to compare electoral tweets from different countries, for example, it will be interesting to determine
% if the distributions of tweets by purpose differ across developed and developing world.
We are also interested in using purpose-annotated tweets as input in a system that automatically summarizes political tweets.
Finally, we hope that a better understanding of purpose of tweets will help drive the political discourse
towards issues and concerns most relevant to the people.

% \section*{Acknowledgments}

\bibliography{references}

\end{document}